\pdfoutput=1

\documentclass[11pt]{article}

\usepackage[final]{acl}

\usepackage{times}
\usepackage{latexsym}

\usepackage[T1]{fontenc}

\usepackage[utf8]{inputenc}

\usepackage{microtype}

\usepackage{inconsolata}

\usepackage{graphicx}
\usepackage{xcolor}
\definecolor{darkgreen}{RGB}{0,100,0}

\usepackage{multirow}
\usepackage[most]{tcolorbox}

\newtcolorbox{previousbox}{
  colback=yellow!20,
  colframe=blue!50!black,
  boxrule=0.5pt,
  arc=1mm,
  left=2pt,
  right=2pt,
  top=2pt,
  bottom=2pt,
  enhanced,
  breakable 
}

\title{STARE at the Structure: Steering ICL Exemplar Selection with Structural Alignment}

\author{First Author \\
  Affiliation / Address line 1 \\
  Affiliation / Address line 2 \\
  Affiliation / Address line 3 \\
  \texttt{email@domain} \\\And
  Second Author \\
  Affiliation / Address line 1 \\
  Affiliation / Address line 2 \\
  Affiliation / Address line 3 \\
  \texttt{email@domain} \\}

\author{
  Jiaqian Li\textsuperscript{1} \quad
  Qisheng Hu\textsuperscript{1} \quad
  Jing Li\textsuperscript{2} \quad
  Wenya Wang\textsuperscript{1} \\
  \textsuperscript{1}Nanyang Technological University, Singapore \\
  \textsuperscript{2}Harbin Institute of Technology, Shenzhen, China \\
  \texttt{m210055@e.ntu.edu.sg}
}

\begin{document}
\maketitle
\begin{abstract}
In-Context Learning (ICL) has become a powerful paradigm that enables LLMs to perform a wide range of tasks without task-specific fine-tuning. However, the effectiveness of ICL heavily depends on the quality of exemplar selection. In particular, for structured prediction tasks such as semantic parsing, existing ICL selection strategies often overlook structural alignment, leading to suboptimal performance and poor generalization. To address this issue, we propose a novel two-stage exemplar selection strategy that achieves a strong balance between efficiency, generalizability, and performance. First, we fine-tune a BERT-based retriever using structure-aware supervision, guiding it to select exemplars that are both semantically relevant and structurally aligned. Then, we enhance the retriever with a plug-in module, which amplifies syntactically meaningful information in the hidden representations. This plug-in is model-agnostic, requires minimal overhead, and can be seamlessly integrated into existing pipelines. Experiments on four benchmarks spanning three semantic parsing tasks demonstrate that our method consistently outperforms existing baselines with multiple recent LLMs as inference-time models. \footnote{Code will be released at \url{https://github.com/Lijiaqian1/ICL-STARE.git}}

\end{abstract}

\section{Introduction}
\begin{figure}[t]
    \centering
    
    \includegraphics[width=\linewidth]{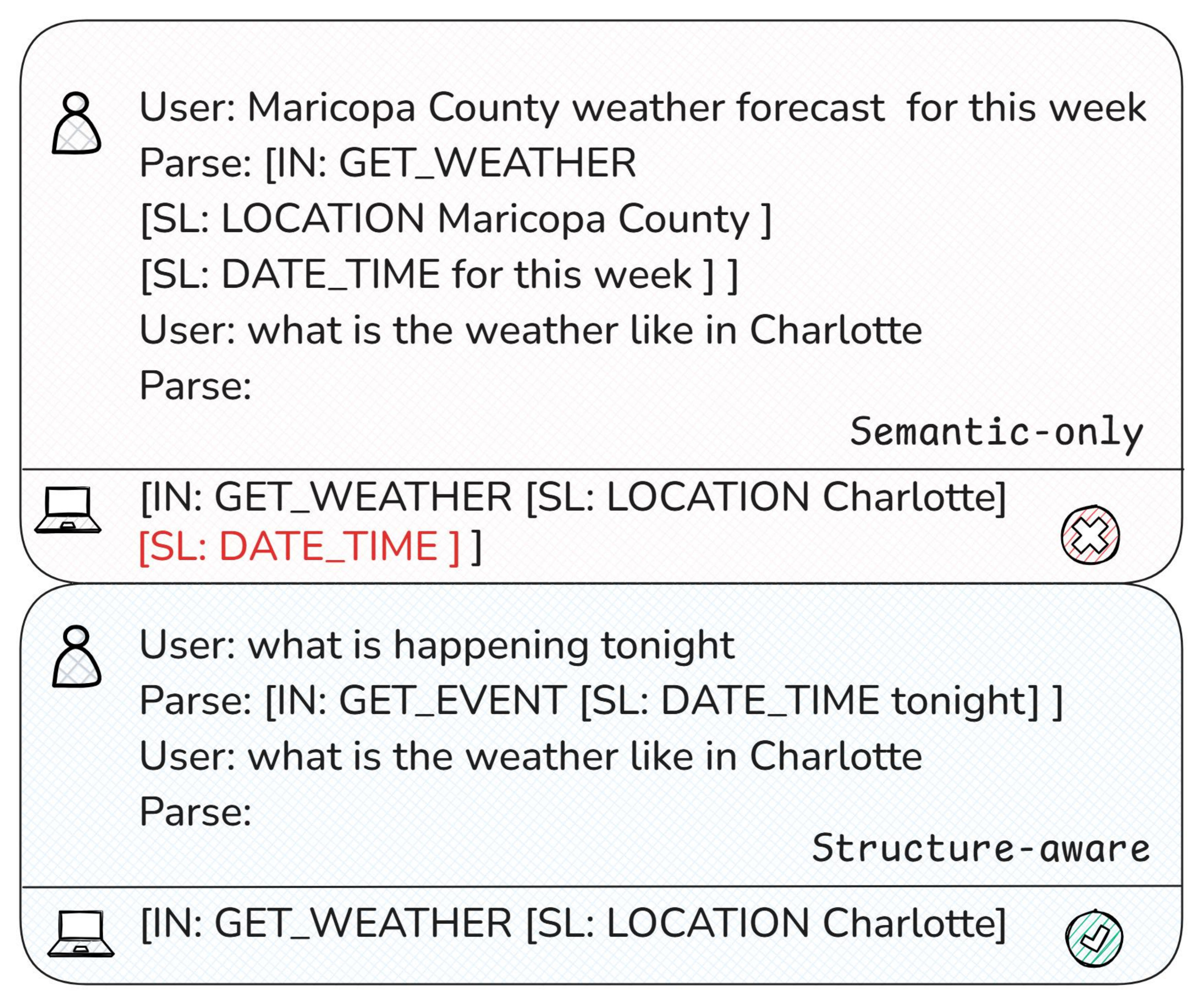}
    \caption{Comparison between semantic-only and structure-aware similarity based one-shot prompting with Llama-3-8B. }
    \label{fig:icl_example}
\end{figure}

Large language models (LLMs) have demonstrated remarkable few-shot capabilities by leveraging in-context learning (ICL) to perform a new task without parameter updates \citep{brown2020language}. 
Despite its effectiveness, prior work has shown that ICL performance is highly sensitive to the choice of exemplars \citep{liu-etal-2022-makes,rubin-etal-2022-learning,li-qiu-2023-finding,li2025taco}. Therefore, how to select meaningful exemplars becomes an active research area.  

Exemplar selection methods can be broadly categorized into two aspects: proxy-task-based \citep{rubin-etal-2022-learning,shi-etal-2022-xricl,li-etal-2023-unified,ye2023compositionalexemplarsincontextlearning} and similarity-based approaches \citep{das-etal-2021-case,hu-etal-2022-context,an-etal-2023-skill}. Proxy-task-based methods extensively query a proxy LLM to evaluate exemplar effectiveness. While effective, they tend to be computationally expensive and often lack generalizability across different models. Similarity-based methods, by contrast, rely on embedding-based metrics to select exemplars that closely match the query instance. However, they typically neglect essential structural information required for precise compositional generalization.

As a result, existing strategies are suboptimal for structure-intensive tasks such as semantic parsing, which involves translating natural language utterances into structured, machine-executable forms, such as logical queries or database commands \citep{zelle1996learning}. These tasks require not only semantic coherence but also precise structural compatibility between exemplars and queries, as illustrated in Figure~\ref{fig:icl_example}.

Meanwhile, many existing exemplar selection methods implicitly rely on the assumption that the model’s learned representations are sufficient for assessing exemplar utility. However, recent interpretability studies suggest that LLM hidden states often encode richer, task-relevant signals than what is directly expressed in their outputs, revealing a gap between internal model knowledge and observable behavior \citep{wang2020languagemodelsopenknowledge,kadavath2022languagemodelsmostlyknow,burns2024discoveringlatentknowledgelanguage}. 

To address these gaps, we propose \textbf{ST}ructure-\textbf{A}ware \textbf{R}etrieval of \textbf{E}xemplars (STARE), a retrieval framework designed for semantic parsing under the ICL paradigm. The framework comprises two key components: 1) a structure-aware retriever that jointly captures both semantic and structural characteristics, and 2) a lightweight plug-in module, Middle-Layer Injection (MLI), that enhances hidden representations with syntactically informative directions. MLI uses linguistic probes and singular value decomposition to identify and amplify syntactic and structural properties in intermediate layers, thereby enhancing the quality of exemplar retrieval. Additionally, the modular design of MLI allows it to be integrated with existing few-shot retrievers, which enables it to enhance semantic parsing performance across diverse scenarios.

To the best of our knowledge, few prior methods for semantic parsing under ICL explicitly incorporate linguistic structure into the retriever’s representations to guide exemplar selection. Experimental results on four diverse semantic parsing benchmarks demonstrate that STARE consistently outperforms existing proxy-based and similarity-based methods, while maintaining lower training costs and exhibiting strong generalizability.

Our contributions can be summarized as follows:

\begin{itemize}
\item We propose STARE, a structure-aware exemplar selection framework for semantic parsing that integrates both semantic and structural criteria.
\item We introduce Middle-Layer Injection (MLI), a lightweight, modular, and model-agnostic technique to enhance hidden representations for improved retrieval.
\item The modular design of MLI allows easy integration with diverse retrieval frameworks, thereby improving the generalizability across tasks and models.
\item Extensive experiments across four benchmarks demonstrate strong performance with lower training costs compared to proxy-task-based methods.

\end{itemize}

\begin{figure*}[t]
    \centering
    \includegraphics[width=\linewidth]{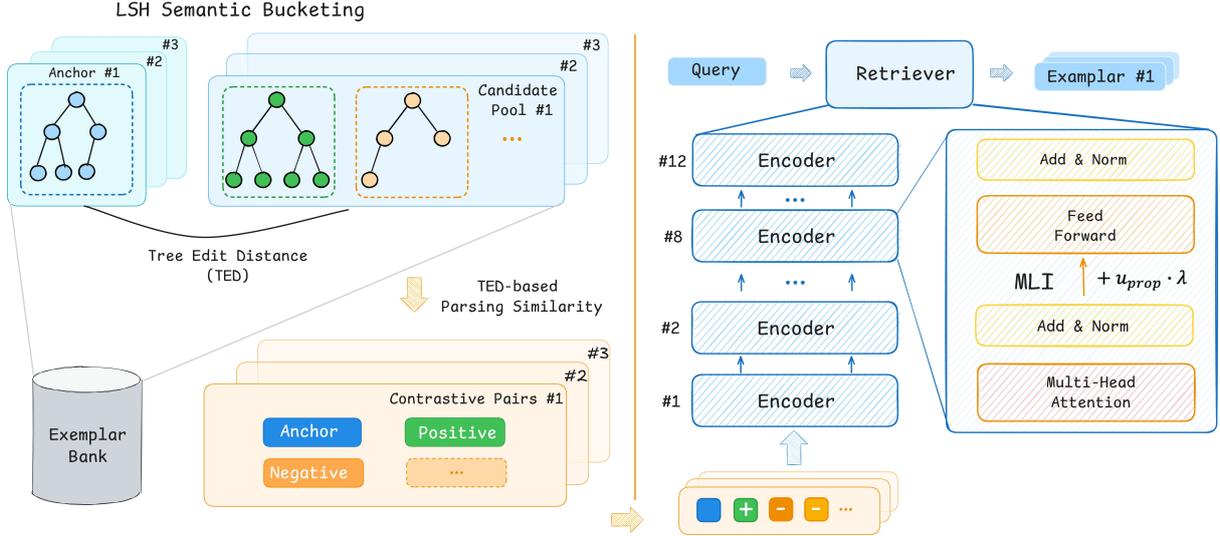}
    \caption{Overview of our proposed framework STARE. The backbone retriever is trained via contrastive learning using semantic and structural similarity signals. The MLI module injects linguistic directions into intermediate hidden states to enhance syntactic awareness. }
    \label{fig:architecture}
\end{figure*}

\section{Related Work}

\paragraph{ICL for Semantic Parsing} 

Early work on semantic parsing with pre-trained models relied on encoder-decoder architectures augmented with schema-aware modules or constrained decoding to ensure well-formed outputs \citep{lin2020bridgingtextualtabulardata, scholak-etal-2021-picard, qi-etal-2022-rasat}. With the advent of stronger models, ICL-based methods emerged. \citet{shin-etal-2021-constrained} showed that few-shot prompting with controlled rephrasings could guide models toward canonical forms before parsing, while \citet{pasupat-etal-2021-controllable} introduced retrieval-augmented ICL. Later, \citet{shin-van-durme-2022-shot} demonstrated that instruction-tuned models can perform direct mappings from natural language to structured forms, shifting the research focus toward improving exemplar selection within ICL setups.

\paragraph{Exemplar Selection for ICL}
Exemplar selection for ICL generally falls into two categories: unsupervised similarity-based and supervised learning-based methods. Unsupervised methods rely on predefined similarity metrics or static retrieval models without task-specific supervision. Common approaches used BM25 or sentence encoders like SBERT to compute semantic similarity between queries and candidate exemplars \citep{liu-etal-2022-makes,agrawal-etal-2023-context}. Skill-KNN enhanced this by extracting task-relevant features to identify skill-overlapping exemplars \citep{an-etal-2023-skill}. TST \citep{poesia2022synchromeshreliablecodegeneration} used the tree edit distance to determine similarities between the query and the candidates for code generation. \citet{wu2023selfadaptiveincontextlearninginformation} proposed a self-adaptive selection framework minimizing entropy under the MDL principle. Supervised methods use explicit training signals. Some defined task-specific metrics such as logical-form alignment \citep{das-etal-2021-case}, slot transitions \citep{hu-etal-2022-context}, or sparse SQL keyword encoding \citep{nan-etal-2023-enhancing}. Others trained retrievers with proxy LLM feedback: Efficient Prompt Retrieval (EPR) used binary labels \citep{rubin-etal-2022-learning}, while Unified Demonstration Retriever (UDR) incorporated ranking and hard negative mining \citep{li-etal-2023-unified}. \citet{ye2023compositionalexemplarsincontextlearning} modeled exemplar selection as subset selection via Determinantal Point Processes (DPPs).

\paragraph{Probing and Representation Intervention}
Probing is a common technique in LLM interpretability that trains diagnostic classifiers on hidden states to identify encoded linguistic properties and analyze their effects on generation \citep{adi2017finegrainedanalysissentenceembeddings,conneau-etal-2018-cram,liu-etal-2019-linguistic}. A more advanced form, causal probing, intervenes in hidden representations to create counterfactuals and assess causal influence \citep{elazar-etal-2021-amnesic,ravfogel-etal-2021-counterfactual}. While initially designed for analysis, such techniques have increasingly been repurposed to steer model behavior. Interventions on hidden states can affect grammatical agreement \citep{tucker-etal-2021-modified}, reduce bias \citep{levy2023probingneedindicatortasks}, enable semantic manipulations via vector arithmetic \citep{subramani-etal-2022-extracting}, and even internalize multimodal ICL \citep{li2025m2ivefficientfinegrainedmultimodal}.
\citet{li2024inferencetimeinterventionelicitingtruthful} introduced Inference-Time Intervention (ITI), which adjusts attention head activations to promote truthful generation. These findings underscore the potential of hidden-state interventions as a powerful tool for behavior control.

\section{Method}
In this section, we introduce the overall methodology of our proposed framework, \textbf{ST}ructure-\textbf{A}ware \textbf{R}etrieval of \textbf{E}xemplars (STARE). The overview of STARE is illustrated in Figure~\ref{fig:architecture}. We begin by formulating the task in Section~\ref{sec:formulation}. Section~\ref{sec:backbone} describes the backbone component of our framework, a finetuned retriever that jointly models semantic and structural similarity. Section~\ref{sec:mli} then introduces Middle-Layer Injection (MLI), a module that enhances the retriever’s syntactic sensitivity by modifying internal representations.

\subsection{Task Formulation}
\label{sec:formulation}
ICL enables LLMs to perform semantic parsing by conditioning on a set of exemplars $\mathcal{E} = \{(x_i, y_i)\}_{i=1}^{k}$, where each $x_i$ is an input query and $y_i$ its corresponding gold parse. Given a test input $x_{\text{test}}$, the model predicts an output $\hat{y}_{\text{test}}$ by maximizing the conditional probability:

\begin{equation}
    \hat{y}_{\text{test}} = \arg\max_{y} P(y \mid x_{\text{test}}, \mathcal{E}; \theta),
\end{equation}

where $\theta$ denotes the frozen model parameters. Since ICL performance is sensitive to the exemplar set $\mathcal{E}$, our goal is to optimize its selection.

We aim to construct an effective retriever $\phi$ that captures both semantic similarity and structural alignment. For each test query $x_{\text{test}}$ and candidate $x_i$ from the training set $\mathcal{D}_{\text{train}}$, we compute embeddings $\phi(x_{\text{test}}), \phi(x_i) \in \mathbf{R}^d$, and select the top-$k$ exemplars based on cosine similarity:

\begin{equation}
    \mathcal{E} = \operatorname{TopK} \big( \cos(\phi(x_{\text{test}}), \phi(x_i)) \big).
\end{equation}

\subsection{Structure-Aware Retriever}
\label{sec:backbone}
In semantic parsing, both semantic context and structural form carry useful signals for exemplar selection. Semantically related exemplars help an LLM recall the appropriate domain knowledge and surface realizations of a parse, whereas structural correspondence provides the most direct guidance for generating a correct and executable output. Consequently, a two-stage strategy is adopted to construct contrastive pairs for retriever training: a coarse semantic bucketing step first collects a high-recall pool of candidates semantically relevant to the anchor exemplar, followed by an evaluation to distinguish structurally aligned exemplars from misaligned ones within the pool.

\subsubsection{Semantic Bucketing}
We compute semantic similarity between parsed outputs rather than input utterances, as logical forms and SQL queries more directly reflect compositional meaning. Since conventional off-the-shelf encoders are poorly suited to formal representations, a hashing-based strategy is adopted to group parses into semantically similar candidate pools.

In practice, each parse $x$ is converted into a set of discrete features (e.g., normalized tokens, keywords, argument labels), denoted as $F(x)$, from which compact MinHash sketches \citep{broder1997resemblance} are generated, providing efficient and order-invariant approximations of Jaccard similarity. These signatures are stored in a Locality-Sensitive Hashing (LSH) index that enables sublinear retrieval of high-similarity candidates by hashing into multiple overlapping buckets. At query time, the anchor parse’s signature is looked up in the LSH index to retrieve all parses with high approximate Jaccard similarity, avoiding exhaustive comparisons against the entire training set.

The LSH index is parameterized by a similarity threshold \( \tau \), which defines the target Jaccard similarity above which two parses are likely to collide in at least one bucket. Concretely, given a training instance's semantic parse \( p \) and its corresponding feature set \( F(p) \), we aim to retrieve \( q \) such that 

\begin{equation}
\mathrm{Jaccard}(F(p), F(q)) = \frac{|F(p) \cap F(q)|}{|F(p) \cup F(q)|} \ge \tau.
\end{equation}

This yields a high-recall candidate pool $\mathcal{C}_p=\{q | \textrm{Jaccard}(F(p), F(q))\geq \tau\}$ for each parse $p$ that is much smaller than the full dataset, and serves as a strong base for contrastive pair construction in the next stage.

\subsubsection{Structure-Based Pair Filtering}
Building on the candidate pool, contrastive pairs are next extracted by measuring structural correspondence between parses.  The goal is to quantify how closely two compositional representations align in their tree topology.

To this end, each semantic parse is converted into a labeled tree and the normalized Zhang–Shasha tree edit distance (TED) \citep{zhang1989simple} is computed. The way to construct tree structure for semantic parses is detailed in Appendix~\ref{appendix:tree_construction}.

Formally, let \( T(p) \) and \( T(q) \) be the trees for parses \( p \) and \( q \). We compute $\mathrm{TED}(T(p), T(q))$
using unit edit costs. This distance is then normalized and converted into a similarity score:

\begin{equation}
\mathrm{sim}_{\mathrm{struct}}(p,q)
= 1 - \frac{\mathrm{TED}(T(p),T(q))}{\max\bigl(|T(p)|, |T(q)|\bigr)} \in [0, 1]
\label{eq:simstruct}
\end{equation}
Higher values indicate closer structural alignment.

For each anchor $p$ with its corresponding candidate pool $\mathcal{C}_p$, the candidate with the highest $\mathrm{sim}_{\mathrm{struct}}$ to the anchor is designated as the positive example, while the least structurally similar candidates within the pool are selected as hard negatives. Additional negatives are randomly sampled from outside the candidate pool. This allows a combination of structure-aware positives and progressively challenging negatives for the contrastive pairs collected.

\subsubsection{Training}
With the structure-aware contrastive pairs, a BERT is fine-tuned as the exemplar retriever. The model is optimized to bring structurally aligned exemplar–query pairs closer in the representation space, while pushing apart structurally divergent or semantically irrelevant ones. We adopt a contrastive learning objective based on the InfoNCE loss \citep{oord2019representationlearningcontrastivepredictive}, where each anchor is paired with one positive and multiple negatives. Sentence-level representations are obtained via mean pooling over the final hidden states of the encoder. 
\subsection{Middle-Layer Injection (MLI)}
\label{sec:mli}
Recent studies have shown that certain
knowledge and properties tend to be attenuated or forgotten as representations progress
through deeper layers \citep{wallat2021bertnesiainvestigatingcaptureforgetting}. This raises a critical challenge in exemplar retrieval: while the retriever is fine-tuned using contrastive learning to align with a predefined similarity metric, it is not guaranteed that the final-layer representations optimally encode the most informative signals for exemplar selection.

Probing techniques offer a diagnostic tool for uncovering what features are encoded in intermediate representations, typically by training linear classifiers to predict certain properties from hidden states \citep{tenney2019bertrediscoversclassicalnlp,hewitt-manning-2019-structural}.  Prior work has explored using probing to identify task-relevant directions in the latent space, for example, directions associated with truthfulness or gender sensitivity. By intervening along these directions, either reinforcing or suppressing them, researchers have been able to increase a model’s likelihood of generating truthful responses or reduce biased behavior \citep{levy2023probingneedindicatortasks,li2024inferencetimeinterventionelicitingtruthful}.

\begin{figure}[t]
    \centering
    \includegraphics[width=\linewidth]{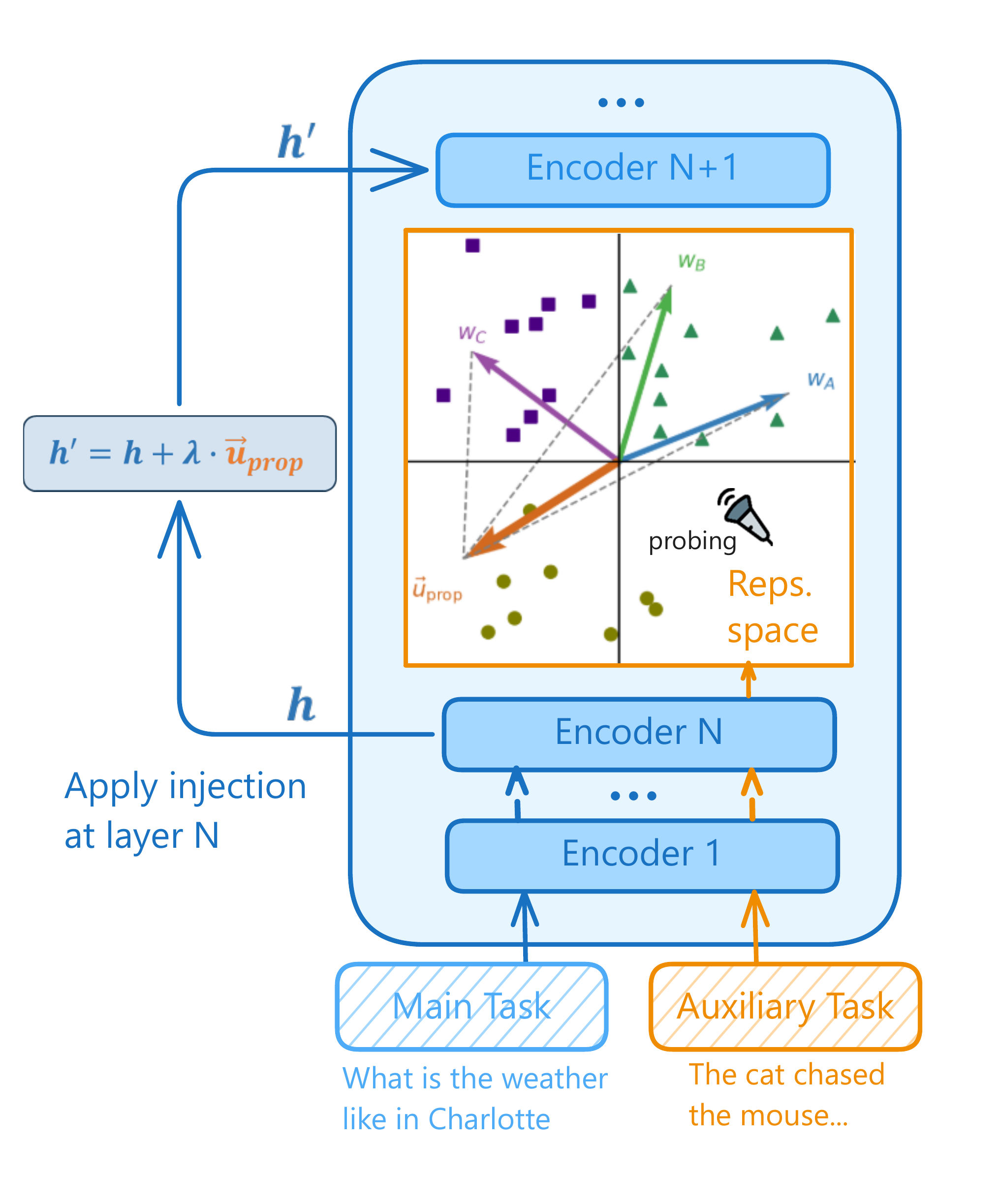}
    \caption{Illustration of Middle-Layer Injection (MLI). The vectors $W_A$, $W_B$, and $W_C$ are rows of the probe weight matrix $W$. $\vec{u}_{\text{prop}}$ is the principal direction extracted from $W$ via SVD.}
    \label{fig:mli}
\end{figure}

Inspired by this, we introduce Middle-Layer Injection (MLI), a method that intervenes in the internal representations of the retriever to amplify task-relevant linguistic abstractions, as illustrated in Figure~\ref{fig:mli}. In the absence of ground-truth utility labels for exemplars, we instead extract directions in the model’s latent space corresponding to well-established linguistic properties and inject these directions into intermediate layers. By enhancing the retriever’s internal encoding of syntactic distinctions, MLI improves the alignment between latent representations and linguistic structure, ultimately leading to more effective exemplar selection.

Concretely, we focus on three widely studied linguistic properties: 
1) \textbf{Part-of-Speech (POS) Tags}, which identify the syntactic category of each token (e.g., noun, verb), 2) \textbf{Dependency Labels (DEPS)}, which define grammatical relationships between words (e.g., subject, object), and 3) \textbf{Phrase Types (PT)}, which describe constituent structures (e.g., noun phrase, verb phrase). 
These properties are chosen because they span fine-grained lexical roles (POS), functional relations (DEPS), and higher-order syntactic structure (PT). Together, they reflect multiple levels of compositional meaning in natural language, making them particularly suitable for enhancing representations used in semantic parsing.

To extract directional signals, auxiliary datasets annotated with the relevant labels are leveraged to train linear probes (logistic regression classifiers) on hidden representations at a chosen layer $N$. Denote the representation for token $w$ at layer $N$ as $h_N \in \mathbf{R}^d$, a linear probe $f_{\text{task}}$ for $\text{task} \in \{\text{POS}, \text{DEPS}, \text{PT}\}$ is obtained by minimizing $L(y, \hat{y})$, where $\hat{y}=f_{task}(h_N)=W_{task}h_N$ and $y$ is the ground-truth label for token $w$. $L$ is the cross-entropy loss function. After training, we denote $\hat{W}_{task} \in \mathbf{R}^{k \times d}$ as the final weight matrix of the classifier, where $k$ is the number of labels.

Singular Value Decomposition (SVD) is then performed on $\hat{W}_{task}$: $\hat{W}_{task} = U \Sigma V^\top$. The first right singular vector $V_1$ from $V$ is selected as the dominant direction $u_{\text{prop}}$ encoding the linguistic property: $ u_{\text{prop}} = V_1$.

To amplify this information in the model, the direction $u_{\text{prop}}$ is injected at the $N$-th layer of the retriever by adjusting each token’s hidden representation $h$:
\begin{equation}
    h' = h + \lambda \, u_{\text{prop}}
\end{equation}
where $\lambda$ controls the intensity of the injection.

The injection layer $N$, augmentation task (POS, DEPS, or PT), and intensity $\lambda$ are hyperparameters. The best configuration is selected by evaluating different combinations on a development set, ensuring that the enhancement provides tangible benefits to retrieval performance.

\begin{table*}[t]
  \centering
  \resizebox{\textwidth}{!}{ 
  \begin{tabular}{lccc|ccc|ccc}
    \hline
    \multirow{2}{*}{\textbf{Method}} 
      & \multicolumn{3}{c|}{\textbf{MTOP}} 
      & \multicolumn{3}{c|}{\textbf{SMCalFlow}} 
      & \multicolumn{3}{c}{\textbf{TreeDST}} \\
    \cline{2-10}
      & \textbf{Llama3} & \textbf{4o-mini} & \textbf{DS-V3} 
      & \textbf{Llama3} & \textbf{4o-mini} & \textbf{DS-V3} 
      & \textbf{Llama3} & \textbf{4o-mini} & \textbf{DS-V3} \\
    \hline
    BM25  & 60.6 & 55.0 & 72.2 & 82.6 & 70.9 & 61.9 & 58.1 & 42.2 & 34.4 \\
    BERT  & 60.5 & 53.3 & 73.9 & 82.0 & 73.2 & 61.9 & 60.1 & 45.4 & 33.9 \\
    MLSM  & 63.6 & 56.9 & 73.3 & 83.0 & 74.5 & 61.0 & 58.1 & 41.9 & 33.5 \\
    EPR   & 67.0 & 58.3 & 73.3 & 82.9 & 73.5 & 63.5 & 60.3 & \textbf{45.5} & 36.1 \\
    CEIL  & 68.3   & 58.8   & 75.7   & 84.3   & 73.8   & \textbf{63.7}   & 60.8   & 44.7   & 35.9   \\
    \textbf{STARE}   & \textbf{69.5} & \textbf{59.4} & \textbf{78.8} 
          & \textbf{86.9} & \textbf{74.8} & 61.9 
          & \textbf{62.1} & \textbf{45.5} & \textbf{37.1} \\
    \hline
  \end{tabular}
  }
  \caption{Exact Match accuracy on MTop, SMCalFlow and TreeDST across different exemplar retrievers and inference models.}
  \label{tab:mst_results}
\end{table*}

\begin{table}[t]
\centering
\small
\setlength{\tabcolsep}{3pt}
\begin{tabular}{lcccccc}
\hline
\textbf{Method} 
& \multicolumn{2}{c}{\textbf{3.5-turbo}} 
& \multicolumn{2}{c}{\textbf{4o-mini}} 
& \multicolumn{2}{c}{\textbf{DS-V3}} \\
& EX & EM & EX & EM & EX & EM \\
\hline
Zero Shot               & 71.2 & 12.1 & 73.0 & 19.1 & 77.0 & 11.7 \\
Random                  & 73.8 & 38.6 & 74.6 & 50.1 & 81.7 & 60.4 \\
BERT                    & 75.5 & 54.4 & 74.2 & 58.4 & 82.3 & 70.5 \\
EPR                     & 73.4 & 48.4 & 74.7 & 55.9 & 82.5 & 67.0 \\
CEIL                    & 75.9 & 42.0 & 74.8 & 50.4 & 82.4 & 66.4 \\
TST                    & 75.1 & 41.3 & 75.0 & 52.4 & 82.7 & 70.9 \\
MLSM                    & 75.3 & 41.6 & 75.2 & 59.9 & 83.4 & 69.4 \\
Skill-KNN (cons.)       & 76.3 & 42.6 & 75.4 & 50.3 & 82.5 & 66.3 \\
Skill-KNN (dist.)       & 76.8 & 43.0 & 72.9 & 49.1 & 82.2 & 63.2 \\
Similarity-Div.         & 75.1 & 42.8 & 76.2 & 51.0 & 82.7 & 65.3 \\
\textbf{STARE}
                        & \textbf{77.0} & \textbf{60.3} 
                        & \textbf{76.9} & \textbf{65.4} 
                        & \textbf{84.5} & \textbf{74.0} \\
\hline
\end{tabular}
\caption{Execution (EX) and Exact Match (EM) accuracy on the Spider dataset across different exemplar retrievers and inference models.}
\label{tab:spider_results}
\captionsetup{justification=justified}
\end{table}

\section{Experiments}

\subsection{Tasks}
Our method is evaluated across four semantic parsing tasks, which span from intent and slot filling (MTop \citep{li-etal-2021-mtop}), task-oriented dialogue parsing (SMCalFlow \citep{andreas2020task}, TreeDST \citep{cheng2021conversationalsemanticparsingdialog}) and text-to-SQL (Spider\citep{yu-etal-2018-spider}). Following standard practice, the training sets of these datasets are used as exemplar banks. Appendix~\ref{appendix:datasets} provides an overview of data splits and examples along with detailed dataset descriptions.
\subsection{Baselines}
Our method is compared against five recent exemplar selection baselines: Efficient Prompt Retriever (EPR) \citep{rubin-etal-2022-learning}, Compositional Exemplars for In-context Learning (CEIL) \citep{ye2023compositionalexemplarsincontextlearning}, Multi-level Similarity Maximization (MLSM) \citep{liu2024unravelingmechanicslearningbaseddemonstration}, Skill-KNN \citep{an-etal-2023-skill}, and Similarity-Diversity \citep{nan-etal-2023-enhancing}. 

EPR and CEIL utilize proxy tasks that incorporate LLM feedback to assess the utility of exemplars. MLSM aggregates similarity signals across BERT layers as expert representations. Skill-KNN and Similarity-Diversity are tailored for text-to-SQL tasks. In addition, unsupervised retrieval baselines such as BM25 and BERT-based dense retrievers are included. Detailed descriptions and implementation details are provided in Appendix~\ref{appendix:baselines}.

\subsection{Experimental Settings}

\paragraph{Backbone Retriever}
A BERT encoder\footnote{\url{https://huggingface.co/bert-base-uncased}} is fine-tuned with InfoNCE loss (temperature 0.07) for at most three epochs using AdamW.  
For each anchor, we construct one positive pair, three hard negative pairs, and two random negative pairs. 

\paragraph{Middle-Layer Injection}
For the auxiliary datasets, we use the English Universal Dependencies (UD) Treebank \cite{mcdonald2013universal} for part-of-speech (POS) and syntactic dependency (DEPS) labels, and the Penn Treebank \cite{marcus1993building} for constituency parsing (Phrase Type, PT) labels. To mitigate overfitting, fine-grained labels are merged into broader categories; the final label sets are summarized in Table~\ref{tab:linguistic_labels} in Appendix~\ref{appendix:merged_labels}. 
The selected properties and  intensities under different settings are listed in Appendix~\ref{appendix:mli_configs}.

\paragraph{Inference LLMs}
For MTop, SMCalFlow, and TreeDST, Llama3-8B \citep{grattafiori2024llama3herdmodels}, GPT-4o-mini \citep{openai2024gpt4omini}, and DeepSeek-V3 \citep{deepseekai2025deepseekv3technicalreport} are used as inference models. For Spider, the same setting is used except that Llama3-8B is replaced with GPT-3.5-turbo \citep{openai2023gpt35turbo}. Detailed settings are provided in Appendix~\ref{appendix:inferencellm_settings}.

\paragraph{Prompt Construction}
Following existing work, we use 20 exemplars for MTop, 5 for SMCalFlow, 10 for TreeDST, and 5 for Spider, ordered by ascending similarity to the test query. The prompts used for Spider incorporate schema linking in the format proposed by~\citet{nan-etal-2023-enhancing}, and adopt the system prompt from~\citet{lee2025safesqlselfaugmentedincontextlearning}.
Full prompt templates and examples are provided in Appendix~\ref{appendix:prompts}.

\paragraph{Evaluation}
Exact Match (EM) is reported for all tasks, while Execution Accuracy (EX) is additionally reported for Spider, following the official evaluation script\footnote{\url{https://github.com/taoyds/test-suite-sql-eval}.}.

\begin{table*}[t]
  \centering
  \resizebox{\textwidth}{!}{ 
  \begin{tabular}{lccc|ccc|ccc}
    \hline
    \multirow{2}{*}{\textbf{Method}} 
      & \multicolumn{3}{c|}{\textbf{MTOP}} 
      & \multicolumn{3}{c|}{\textbf{SMCalFlow}} 
      & \multicolumn{3}{c}{\textbf{TreeDST}} \\
    \cline{2-10}
      & \textbf{Llama3} & \textbf{4o-mini} & \textbf{DS-V3} 
      & \textbf{Llama3} & \textbf{4o-mini} & \textbf{DS-V3} 
      & \textbf{Llama3} & \textbf{4o-mini} & \textbf{DS-V3} \\
    \hline
    STARE (w/o MLI) & 67.8 & 57.5 & 74.5 & 84.3 & 71.6 & 63.0 & 61.0 & 43.8 & 36.2 \\
    STARE           & \textbf{69.5} & 59.4 & \textbf{78.8} & 86.9 & 74.8 & 61.9 & \textbf{62.1} & 45.5 & 37.1 \\
    $\Delta$ (MLI Gain) & \textcolor{darkgreen}{+1.7} & \textcolor{darkgreen}{+1.9} & \textcolor{darkgreen}{+4.3} & \textcolor{darkgreen}{+2.6} & \textcolor{darkgreen}{+3.2} & \textcolor{red}{-1.1} & \textcolor{darkgreen}{+1.1} & \textcolor{darkgreen}{+1.7} & \textcolor{darkgreen}{+0.9} \\
    \hline
    BERT & 60.5 & 53.3 & 73.9 & 82.0 & 73.2 & 61.9 & 60.1 & 45.4 & 33.9 \\
    BERT + MLI     & 64.4 & 58.7 & 76.2 & \textbf{87.5} & 73.9 & 62.2 & 61.6 & 46.1 & 35.7 \\
    $\Delta$ (MLI Gain) & \textcolor{darkgreen}{+3.9} & \textcolor{darkgreen}{+5.4} & \textcolor{darkgreen}{+2.3} & \textcolor{darkgreen}{+5.5} & \textcolor{darkgreen}{+0.7} & \textcolor{darkgreen}{+0.3} & \textcolor{darkgreen}{+1.5} & \textcolor{darkgreen}{+0.7} & \textcolor{darkgreen}{+1.8} \\
    \hline
    EPR & 67.0 & 58.3 & 73.3 & 82.9 & 73.5 & \textbf{63.5} & 60.3 & 45.5 & 36.1 \\
    EPR + MLI     & 67.4 & \textbf{60.6} & 75.7 & 86.3 & \textbf{75.2} & 62.5 & 60.6 & \textbf{46.3} & \textbf{40.0} \\
    $\Delta$ (MLI Gain) & \textcolor{darkgreen}{+0.4} & \textcolor{darkgreen}{+2.3} & \textcolor{darkgreen}{+2.4} & \textcolor{darkgreen}{+3.4} & \textcolor{darkgreen}{+1.7} & \textcolor{red}{-1.0} & \textcolor{darkgreen}{+0.3} & \textcolor{darkgreen}{+0.8} & \textcolor{darkgreen}{+3.9} \\
    \hline

  \end{tabular}
  }
  \caption{Effect of MLI on Exact Match accuracy across different exemplar retrievers and inference models on MTop, SMCalFlow and TreeDST.}
  \label{tab:plugin_main}
\end{table*}

\begin{table}[t]
\centering
\small
\setlength{\tabcolsep}{3pt} 
\begin{tabular}{lcccccc}
\hline
\textbf{Method} 
& \multicolumn{2}{c}{\textbf{3.5-turbo}} 
& \multicolumn{2}{c}{\textbf{4o-mini}} 
& \multicolumn{2}{c}{\textbf{DS-V3}} \\
& EX & EM & EX & EM & EX & EM \\
\hline
STARE (w/o MLI) & 76.5 & 57.0 & 76.8 & 60.7 & 83.0 & 72.1 \\
STARE           & 77.0 & 60.3 & 76.9 & 65.4 & 84.5 & 74.0 \\
$\Delta$ (MLI Gain) & \textcolor{darkgreen}{+0.5} & \textcolor{darkgreen}{+3.3} & \textcolor{darkgreen}{+0.1} & \textcolor{darkgreen}{+4.7} & \textcolor{darkgreen}{+1.5} & \textcolor{darkgreen}{+1.9} \\
\hline
BERT            & 75.5 & 54.4 & 74.2 & 58.4 & 82.3 & 70.5 \\
BERT + MLI           & 75.6 & 56.3 & 74.3 & 61.1 & 83.9 & 72.0 \\
$\Delta$ (MLI Gain) & \textcolor{darkgreen}{+0.1} & \textcolor{darkgreen}{+1.9} & \textcolor{darkgreen}{+0.1} & \textcolor{darkgreen}{+2.7} & \textcolor{darkgreen}{+1.6} & \textcolor{darkgreen}{+1.5} \\
\hline
EPR             & 73.4 & 48.4 & 74.7 & 55.9 & 82.5 & 67.0 \\
EPR + MLI           & 76.6 & 57.1 & 75.5 & 60.7 & 83.8 & 73.0 \\
$\Delta$ (MLI Gain) & \textcolor{darkgreen}{+3.2} & \textcolor{darkgreen}{+8.7} & \textcolor{darkgreen}{+0.8} & \textcolor{darkgreen}{+4.8} & \textcolor{darkgreen}{+1.3} & \textcolor{darkgreen}{+6.0} \\
\hline

\end{tabular}
\caption{Effect of MLI on Execution (EX) and Exact Match (EM) accuracy across different exemplar retrievers and inference models on the Spider dataset.}
\label{tab:plugin_spider}
\end{table}

\section{Results}

\subsection{Main Results}

We report the main ICL performance of our proposed framework STARE across the four semantic parsing benchmarks. Results for MTop, SMCalFlow, and TreeDST are summarized in Table~\ref{tab:mst_results}, while Table~\ref{tab:spider_results} presents Execution accuracy (EX) and Exact Match (EM) on the Spider dataset. Supplementary results are provided in Appendix~\ref{appendix:supp_results}.

Our method STARE consistently outperforms all baselines, including proxy-task-based methods such as EPR and CEIL, except on SMCalFlow under DeepSeek-V3. On average, STARE yields a 1.35\% gain over the strongest competing baseline across the first three tasks. On Spider, STARE achieves the best EX and EM across all inference models, improving over the best baseline by 0.9\% (EX) and 5.0\% (EM).

In contrast to EPR and CEIL, which depend on intensive LLM interactions to derive training supervision, STARE avoids reliance on the inference model during training. This not only improves efficiency, but also mitigates overfitting to biases introduced by the proxy model, which can compromise generalization to stronger inference models.

\subsection{MLI as a Plug-in}
To assess the contribution of Middle-Layer Injection (MLI) within our STARE framework, we first compare STARE with and without MLI to isolate the effect of linguistic augmentation. Furthermore, to examine the generalizability of MLI beyond our framework, MLI is applied as a plug-in module to two representative baselines: an unsupervised BERT retriever and the supervised Efficient Prompt Retriever (EPR). Results are shown in Table~\ref{tab:plugin_main} (MTop, SMCalFlow, TreeDST) and Table~\ref{tab:plugin_spider} (Spider). MLI configuration details determined based on development set are listed in Appendix~\ref{appendix:mli_configs}. Empirical results show that Spider benefits more from lexical cues (POS), whereas tasks such as SMCalFlow and TreeDST gain more from structural cues (DEPS, PT). This observation is consistent with our design rationale, which emphasizes the inclusion of complementary syntactic properties to better accommodate the diverse requirements of different semantic parsing tasks.

Experimental results across MTop, SMCalFlow, and TreeDST show that MLI enhances STARE performance under all three inference LLMs, with an average improvement of 2.2\%. The only exception is SMCalFlow when using DeepSeek-V3, where performance slightly declines. On the Spider dataset, MLI provides an average gain of 0.7\% in execution accuracy and 3.3\% in exact match.

Notably, MLI also improves retrieval performance when integrated with both BERT and EPR retrievers. In some cases, MLI enables a non-fine-tuned BERT retriever to achieve performance comparable to fully supervised retrievers, for example, on SMCalFlow with Llama3-8B and TreeDST with GPT-4o-mini. On the Spider dataset, MLI achieves up to an 8.7\% boost in exact match when combined with EPR under GPT-3.5-turbo.

These results underscore MLI's versatility as a modular plug-in. It can be applied to both unsupervised and supervised retrievers and is especially useful in low-resource scenarios where retriever training is infeasible.

\begin{figure*}[t]
  \centering
  \includegraphics[width=\linewidth]{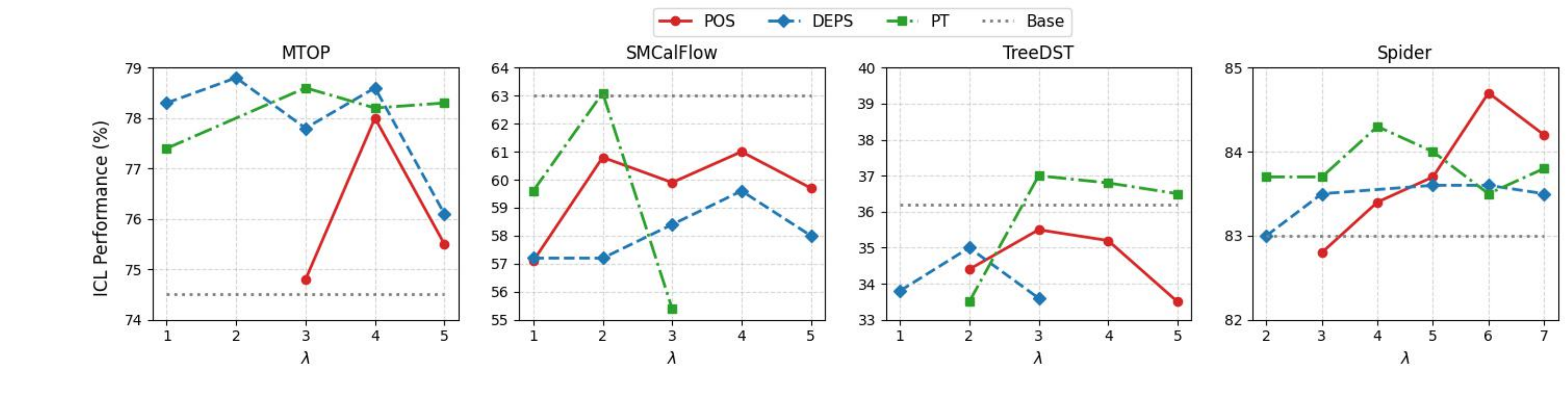}
  \caption{ICL performance on the dev sets of MTop, SMCalFlow, and TreeDST (reported as Exact Match accuracy), and Spider (reported as Execution accuracy), under varying MLI injection intensities \( \lambda \) for each linguistic property, using DeepSeek-V3 as the inference model.}
  \label{fig:mli_variations}
\end{figure*}

\section{Ablation Study}
To investigate how the choice of injection layer, the augmented linguistic property, and the augmentation intensity $\lambda$ affect performance, a series of ablation studies are conducted by systematically varying these factors and analyzing their impact.

\paragraph{Injection Layer}
Injecting MLI into all 12 BERT layers is computationally costly and often redundant due to high inter-layer similarity from residual connections. To reduce overhead, we draw on BERTology findings that different layers capture distinct linguistic patterns: lower layers (1–4) encode lexical features, middle layers (5–9) capture syntax, and upper layers (10–12) model semantics and task-specific abstractions~\citep{ethayarajh-2019-contextual,tenney2019bertrediscoversclassicalnlp,jawahar-etal-2019-bert}. Based on this, Layers 4, 8, and 12 are pre-selected as injection candidates.

\begin{figure}[t]
  \centering
  \includegraphics[width=\linewidth]{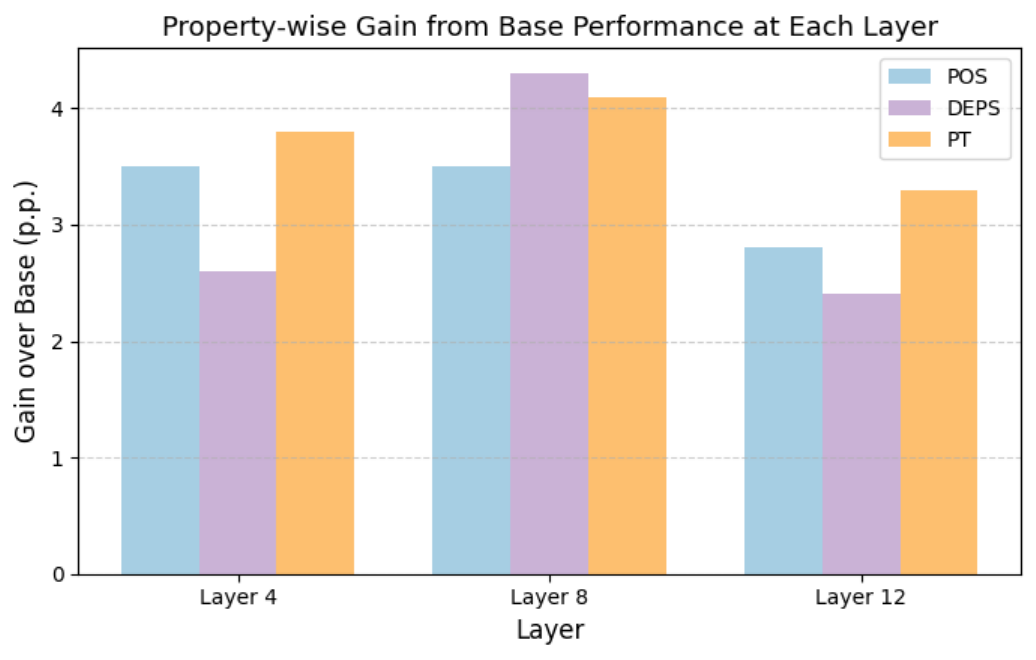}
  
  \caption{Effect of applying MLI at different injection layers on MTop performance, using DeepSeek-V3 as the inference model.}
  \label{fig:mtop_layer}
  
\end{figure}

For each layer, we apply property-specific injections (POS, DEPS, PT) across a range of intensity values ($\lambda = 1, 2, 3, 4, 5$), and record the maximum performance gain. Figure~\ref{fig:mtop_layer} shows the results on MTop using DeepSeek-V3.

The injection at Layer 8 consistently achieves the highest gains across all properties. We attribute this to its intermediary position, allowing injected signals to propagate effectively while still benefiting from well-encoded linguistic features. In contrast, early injection may dilute the signal, and late injection may limit downstream utilization. Further analysis of the injection layer ablation is presented in Appendix~\ref{appendix:injection_layer}.

\paragraph{MLI Configurations}
\label{sec:mli_configs}
To determine the most effective linguistic property for injection and the optimal intensity $\lambda$, controlled experiments are conducted on the development set. For each of the three properties (POS, DEPS, and PT), we sweep over a range of $\lambda$ values and observe the resulting ICL performance. Figure~\ref{fig:mli_variations} illustrates this tuning process when DeepSeek-V3 is used as the inference model\footnote{For optimal presentation, we omit results for those \(\lambda\) falling out of the range in the figure}. The dashed line represents the performance of our backbone retriever without MLI. Typically, we observe that the characteristic trend of contributing properties is an initial increase followed by a decline as $\lambda $ grows.  We select the configuration that achieves the highest development performance for each task and apply it in test-time evaluation. 

The observed trend indicates that there is an optimal injection strength. A small $\lambda$ may fail to meaningfully influence the representation, while overly aggressive injection risks disrupting the structural integrity of the latent space.

\section{Conclusion}
We present Structure-Aware Retrieval of Exemplars (STARE), a novel framework for in-context learning exemplar retrieval  for semantic parsing. STARE combines a backbone retriever based on semantic hashing and dependency tree representations with a modular enhancement strategy, Middle-Layer Injection (MLI). MLI serves as a lightweight yet effective augmentation mechanism, which can be integrated into various retrieval pipelines. Our method achieves state-of-the-art performance across multiple semantic parsing benchmarks, while maintaining low training costs and demonstrating strong generalizability.

\section*{Limitations}
Our Middle-Layer Injection (MLI) method assumes a linear mechanism of injecting linguistic properties, treating each direction independently in the hidden space. This simplification may overlook potential nonlinear interactions, such as those between part-of-speech tags and syntactic dependencies, which may be important in certain retrieval scenarios and warrant further investigation in future work. In addition, our evaluation focuses solely on exemplar selection, without comparisons to reasoning-centric methods such as chain-of-thought prompting or to fine-tuning-based approaches, which may limit the scope of our current analysis to retrieval-specific settings.

\section*{Acknowledgments}
This research is supported by the Ministry of Education, Singapore, under its Academic Research
Fund Tier 1 (\#023618-00001, RG99/23), and the
NTU Start-Up Grant (\#023284-00001), Singapore.

\bibliography{main}

\newpage

\appendix

\section{Datasets}
\label{appendix:datasets}
Table~\ref{tab:datasets} presents the data splits and example instances for each dataset. Detailed descriptions are provided below.

\paragraph{MTop} 
MTop \cite{li-etal-2021-mtop} is a multilingual task-oriented semantic parsing dataset spanning diverse user intents and domains. We focus on the English subset. We use the full training set and sample 1K examples each from the original validation and test sets to form our splits.

\paragraph{SMCalFlow}
SMCalFlow \cite{andreas2020task} is a conversational semantic parsing dataset serialized in a LISP-style format (Lispress). Its structured representation enforces well-formedness and supports generalization in low-data settings, making it ideal for testing compositional generalization in dialogue. We sample 5K/1K/1K examples from its original training, validation and test datasets as our splits.
\paragraph{TreeDST}
TreeDST \cite{cheng2021conversationalsemanticparsingdialog} is a task-oriented dialogue dataset representing dialogue states as hierarchical trees. We use its Lispress serialization version \cite{platanios-etal-2021-value}, which captures compositional dependencies across intents, domains, and slots, better reflecting real-world dialogue complexity. We sample 5K/1K/1K examples from its original training, validation and test datasets as our splits.

\paragraph{Spider}
Spider \cite{yu-etal-2018-spider} is a large-scale text-to-SQL dataset. It covers a wide range of domains and compositional SQL structures, providing a rigorous testbed for text-to-SQL exemplar selection methods. Its different splits do not share any databases. Following standard practice, we evaluate on the development set since the test set is not publicly released.

\begin{table*}
  \centering
  \footnotesize
  \begin{tabular}{lll}
    \hline
    \textbf{Dataset} & \textbf{Train/Dev/Test} & \textbf{Example} \\
    \hline
    MTop & 15,567/1,000/1,000 &
    \begin{tabular}[t]{@{}l@{}}
      \textbf{Sentence:} Whats weather forecast for tomorrow? \\
      \textbf{Parsing:} [IN:GET\_WEATHER [SL:DATE\_TIME for tomorrow]]
    \end{tabular}
    \\
    SMCalFlow & 50,000/1,000/1,000 &
    \begin{tabular}[t]{@{}l@{}}
      \textbf{Sentence:} What does my schedule look like on Thursday? \\
      \textbf{Parsing:} (Yield (FindEventWrapperWithDefaults \\
      \ \ (EventOnDate (NextDOW (Thursday)) (\^{}(Event) EmptyStructConstraint))))
    \end{tabular}
    \\
    TreeDST & 50,000/1,000/1,000 &
    \begin{tabular}[t]{@{}l@{}}
      \textbf{Sentence:} Hi my assistant, where is the Westin hotel? \\
      \textbf{Parsing:} (plan (\^{}(Hotel) Find :focus \\
      \ \ (Hotel.location\_? (\^{}(String) always)) :object (Hotel.hotelName\_? (?= "Westin"))))
    \end{tabular}
    \\
    Spider & 7,000/1,000/1,034 &
    \begin{tabular}[t]{@{}l@{}}
      \textbf{Sentence:} How many available features are there in total? \\
      \textbf{Parsing:} SELECT count(*) FROM Other\_Available\_Features
    \end{tabular}
    \\
    \hline
  \end{tabular}
  \caption{Overview of datasets used for semantic parsing experiments.}
  \label{tab:datasets}
\end{table*}

\section{Baselines}
\label{appendix:baselines}
\paragraph{Efficient Prompt Retriever (EPR)}
EPR \citep{rubin-etal-2022-learning} constructs contrastive training pairs by applying one-shot prompting to every training instance, using a proxy LLM to approximate exemplar utility. Specifically, a simple retriever is first used to retrieve a candidate pool, and top-K and bottom-K candidates are selected based on proxy model scores. A BERT-based retriever is then fine-tuned via contrastive learning. Following the original setup, we use GPT-Neo (2.7B) as the proxy model. We then retrieve 15 candidates for each training instance using BM25, and set $K=5$ for pair construction.

\paragraph{Compositional Exemplars for In-context Learning (CEIL)}
CEIL \citep{ye2023compositionalexemplarsincontextlearning} builds on the EPR framework but improves selection granularity by modeling exemplar interactions. Instead of scoring one-shot prompts individually, it adopts Determinantal Point Processes (DPP) to select exemplar subsets that jointly maximize compositional contribution. As with EPR, we use BM25 for candidate pre-selection, GPT-Neo (2.7B) as the scoring model, and score 10 randomly sampled subsets of 16 examples for each training instance.

\paragraph{Target Similarity Tuning (TST)}
TST \citep{poesia2022synchromeshreliablecodegeneration} selects exemplars for in-context learning in code generation by ranking candidate programs with respect to the query using Tree-Edit Distance (TED). While both TST and STARE exploit structural cues, there are key distinctions. First, TST learns via regression to absolute TED values, whereas STARE employs a contrastive objective that only preserves relative ordering. Second, TST requires $O(N^{2})$ TED computations across all candidate pairs, while STARE leverages semantic bucketing (MinHash+LSH) to achieve $O(N \log N)$ efficiency. Finally, TST was designed for earlier models with frequent syntax errors and thus relied on external syntax correction, whereas STARE is tailored to modern LLMs and directly addresses structural alignment through both explicit (contrastive retriever) and implicit (MLI) mechanisms, without additional decoding modules.

\paragraph{Multi-level Similarity Maximization (MLSM)}
MLSM \citep{liu2024unravelingmechanicslearningbaseddemonstration} proposes to leverage different abstraction levels captured across BERT layers for exemplar selection. Redundant layers are first filtered using CKA-based clustering, and each selected layer acts as an expert capturing similarity at a distinct level. For each query, MLSM aggregates similarity scores from multiple layers with learned weights, optimizing for agreement across experts. The method is fully unsupervised and designed to enhance task-agnostic generalization without relying on task-specific labels.

\paragraph{Skill-KNN}
Skill-KNN \citep{an-etal-2023-skill} proposes a two-stage retrieve framework for text-to-SQL in-context learning. First, for each query, a frozen LLM is prompted to rewrite the input into a skill-based description that captures task-relevant features in natural language. These rewritten skills are then embedded with an off-the-shelf encoder, and examples with similar skills are retrieved. To address the sensitivity of rewriting to prompt order, two variants are proposed: a consistency-based variant, which aggregates multiple rewrites via mean pooling, and a distinctiveness-based variant, which selects based on the most distinctive match. In our experiments, we use GPT-4o-mini to generate 5 candidate skill descriptions per input and evaluate both variants. We use \texttt{bert-base-uncased} as the off-the-shelf encoder for embedding the skill descriptions.

\paragraph{Similarity-Diversity}
Similarity-Diversity \citep{nan-etal-2023-enhancing} proposes selecting exemplars by balancing similarity and diversity among demonstrations. First, candidates are filtered by retrieving examples with similar SQL structure complexity, using the difficulty-level categorization from the Spider dataset. Then, to promote diversity, a sparse encoding of the predicted SQL query is computed, and $k$-means clustering is applied over the discrete representations to select diverse exemplars. In the original setup, an approximate SQL prediction by baseline text-to-SQL models is used for difficulty categorization and sparse encoding. To avoid introducing noise from imperfect preliminary models, we instead use the ground-truth SQL queries for encoding in our experiments, representing an upper-bound variant of this method.

\paragraph{BM25 Retriever}
BM25 \citep{robertson2009probabilistic} is a sparse retrieval baseline that scores exemplar candidates by computing lexical similarity with the test query. Specifically, it compares each candidate’s input utterance to the test query using term frequency and inverse document frequency, and selects the Top-K highest-scoring candidates.

\paragraph{BERT Retriever}
The BERT retriever encodes both exemplars and test queries using a pre-trained BERT model and selects those with the highest cosine similarity in embedding space. Despite being unsupervised, it captures richer semantic signals than token-level matching methods like BM25, and serves as a lightweight neural baseline for retrieval.

\section{Tree Construction}
\label{appendix:tree_construction}

To enable structure-based similarity computation, we convert each semantic parse into a labeled tree.

For MTop, SMCalFlow and TreeDST, we use a bracket-based parser to recursively construct trees, where each non-terminal label becomes a parent node, and its enclosed spans are attached as children. The resulting tree captures the hierarchical structure of the parse.

For SQL Queries (Spider), we parse SQL queries into Abstract Syntax Trees (ASTs) using \texttt{sqlglot}, and prune them by retaining only clause-level nodes and essential fields to form a structural skeleton.

In both cases, each node’s label corresponds to its operator or clause type, and children reflect its compositional arguments.

\section{Final Merged Labels of Auxiliary Datasets}
\label{appendix:merged_labels}
Table ~\ref{tab:linguistic_labels} shows the merged linguistic labels for the three properties in the auxiliary datasets. Specifically, the POS and DEPS labels are derived from the UD Treebank~\citep{mcdonald2013universal}, which contains 207,230 tokens, while the PT labels are sourced from the Penn Treebank~\citep{marcus1993building}, comprising 100,676 tokens.

\begin{table}[t]
\centering
\small
\setlength{\tabcolsep}{4pt}
\renewcommand{\arraystretch}{1.2}
\begin{tabular}{lp{6.2cm}}  
\hline
\textbf{Property} & \textbf{Final Merged Labels} \\
\hline
POS & ADJ, ADP, ADV, AUX, CCONJ, DET, INTJ, NOUN, NUM, PART, PRON, PROPN, PUNCT, SCONJ, SYM, VERB, X \\
DEPS & ACL, ACL:RELCL, ADVMOD, AUX, CASE, COMP, COMPOUND, CONJ, CSUBJ, DEP, DET, EXPL, GOESWITH, LIST, MARK, MOD, NMOD, NSUBJ, OBJ, OBL, ORPHAN, PUNCT, REPARANDUM, ROOT, VOCATIVE \\
PT & SBAR, UCP, ADVP, O, WHADVP, NAC, INTJ, NX, CONJP, QP, SBARQ, S, ADJP, FRAG, SQ, LST, PRT, PP, X-HLN, VP, X, WHADJP, WHPP, NP, WHNP, SINV, PRN \\
\hline
\end{tabular}
\caption{Final label sets used for linguistic probing after merging fine-grained categories.}
\label{tab:linguistic_labels}
\end{table}

\section{Inference Settings}
\label{appendix:inferencellm_settings}
For all experiments, we use greedy decoding, with the temperature set to 0.0 to ensure deterministic generation. The maximum generation length is capped at 200 new tokens, excluding the input prompt. To ensure robustness, each experiment is run with three different random seeds. The final reported results are obtained by averaging over these three runs.

\section{Supplementary Results}
\label{appendix:supp_results}
Table~\ref{tab:supp_llama_mtop} to Table~\ref{tab:supp_qwen_spider} show the supplementary experimental results with Llama3.3-70B and Qwen2.5-72B-Instruct-Turbo as inference models.
\begin{table}[h]
\centering
\footnotesize
\begin{tabular}{lccc}
\hline
\textbf{Method} & \textbf{MTop} & \textbf{SMCalFlow} & \textbf{TreeDST} \\
\hline
BERT            & 71.0 & 56.8 & 50.0 \\
MLSM            & 71.9 & 57.4 & 49.9 \\
EPR             & 72.1 & 61.3 & 55.7 \\
STARE (w/o MLI) & 72.2 & 59.6 & 53.0 \\
\textbf{STARE}  & \textbf{73.2} & \textbf{61.6} & \textbf{57.2} \\
\hline
\end{tabular}
\caption{Supplementary results with \textbf{Llama 3.3-70B} on MTop/SMCalFlow/TreeDST.}
\label{tab:supp_llama_mtop}
\end{table}

\begin{table}[h]
\centering
\footnotesize
\begin{tabular}{lcc}
\hline
\textbf{Method} & \textbf{EM} & \textbf{EX} \\
\hline
BERT              & 67.0 & 79.5 \\
MLSM              & 66.9 & 80.2 \\
EPR               & 63.2 & 77.2 \\
Skill-KNN (cons.) & 60.3 & 78.9 \\
Skill-KNN (dist.) & 60.2 & 80.6 \\
Similarity-Div.   & 58.4 & 80.4 \\
STARE (w/o MLI)   & 69.5 & 80.9 \\
\textbf{STARE}    & \textbf{72.7} & \textbf{82.1} \\
\hline
\end{tabular}
\caption{Supplementary results with \textbf{Llama 3.3-70B} on Spider.}
\label{tab:supp_llama_spider}
\end{table}

\begin{table}[h]
\centering
\footnotesize
\begin{tabular}{lccc}
\hline
\textbf{Method} & \textbf{MTop} & \textbf{SMCalFlow} & \textbf{TreeDST} \\
\hline
BERT            & 73.2 & 89.1 & 62.7 \\
MLSM            & 72.1 & 89.9 & 60.8 \\
EPR             & 72.2 & 87.9 & \textbf{65.1} \\
STARE (w/o MLI) & 72.6 & 89.6 & 64.1 \\
\textbf{STARE}  & \textbf{74.6} & \textbf{91.2} & 64.8 \\
\hline
\end{tabular}
\caption{Supplementary results with \textbf{Qwen2.5-72B-Instruct-Turbo} on MTop/SMCalFlow/TreeDST.}
\label{tab:supp_qwen_mtop}
\end{table}

\begin{table}[h]
\centering
\footnotesize
\begin{tabular}{lcc}
\hline
\textbf{Method} & \textbf{EM} & \textbf{EX} \\
\hline
BERT              & 65.6 & 80.8 \\
MLSM              & 66.2 & 80.4 \\
EPR               & 64.0 & 81.3 \\
Skill-KNN (cons.) & 62.7 & 81.3 \\
Skill-KNN (dist.) & 60.9 & 80.6 \\
Similarity-Div.   & 62.0 & 81.5 \\
STARE (w/o MLI)   & 71.2 & 82.1 \\
\textbf{STARE}    & \textbf{71.4} & \textbf{82.8} \\
\hline
\end{tabular}
\caption{Supplementary results with \textbf{Qwen2.5-72B-Instruct-Turbo} on Spider.}
\label{tab:supp_qwen_spider}
\end{table}

\section{Injection Layer}
\label{appendix:injection_layer}
To explain why POS, DEPS, and PT injections exhibit different effect trends across layers, we draw insights from BERT interpretability studies and consider each property’s representational footprint in Transformer encoders.

POS tags are encoded early in the network. Empirical probing shows that POS linear separability is high at Layer 4 of BERT-base \citep{tenney2019bertrediscoversclassicalnlp}, so Layer 4 offers a clear axis for lexical separation. Nudging along this axis therefore improves retrieval. Layer 8 still retains strong lexical cues while adding richer context (e.g., clause boundaries, mid-range dependencies). Injecting the POS direction here sharpens lexical roles further and allows downstream layers to integrate this with syntactic context, so the gain remains comparable to Layer 4. Beyond Layer 10, the encoder focuses more on sentence-level semantics and pools token details more aggressively, so lexical perturbations are increasingly diluted, which explains the drop at Layer 12.

In contrast, DEPS and PT emerge only after self-attention has aggregated sufficient context (around Layers 6–9). Injecting a dependency- or phrase-type direction too early pushes the representation along a subspace that has not yet been reliably formed, which actually adds noise. Layer 8 sits at a stage where syntactic structure is both clear and still flexible, so the model can amplify and propagate the signal, yielding the largest gains. By Layer 12, high-level semantics dominate and such perturbations are mostly washed out, so the benefit diminishes.

This pattern is consistent with the structural-probe findings of \citet{hewitt-manning-2019-structural}, which showed that BERT’s middle layers best encode syntactic structure for reconstructing gold dependency trees. In their experiments, both the Undirected Unlabeled Attachment Score (UUAS) and the Spearman correlation between true and predicted parse distances peak at Layer 8, confirming that this layer encodes syntactic geometry most faithfully and is therefore the most effective anchor for our DEPS and PT intervention.

Finally, it is shown that Layer 4 still outperforms Layer 12, albeit by a smaller margin, because early layers retain fine-grained positional information and short-range head–dependent cues that are useful once amplified. Late layers, in contrast, have already compressed many token-level distinctions in favor of sentence-level semantics; injecting syntactic signals there offers little additional discriminatory power for our retriever.

Thus, the observed hierarchy naturally follows from BERT’s progressive shift in representational focus from lexical to syntactic and then to semantic information as depth increases.

\newpage
\onecolumn
\section{Prompt Examples}
\label{appendix:prompts}

\subsection{Prompt Example for MTop, SMCalFlow, TreeDST}
\label{appendix:prompt_mtop}

\noindent
(Same prompt template is used for MTop, SMCalFlow and TreeDST while the following example is instantiated with MTop) \\
\noindent
\\Below are examples of converting user utterances into MTop semantic parses:

\bigskip

\textbf{Example 1}\\
User: Remind me about shopping for school on tax free weekend.\\
Parse: [IN:CREATE\_REMINDER [SL:PERSON\_REMINDED me ] [SL:TODO shopping for school ] [SL:DATE\_TIME on tax free weekend ] ]

\bigskip

\textbf{Example 2}\\
User: Remind me to bake cookies tomorrow night for the bake sale\\
Parse: [IN:CREATE\_REMINDER [SL:PERSON\_REMINDED me ] [SL:TODO [IN:GET\_TODO [SL:TODO bake cookies ] [SL:DATE\_TIME tomorrow night ] [SL:TODO the bake sale ] ] ] ]

\bigskip

\textbf{Example 3}\\
User: Remind me to tell Angie I am bringing the salad for bible study on Friday\\
Parse: [IN:CREATE\_REMINDER [SL:PERSON\_REMINDED me ] [SL:TODO tell Angie I am bringing the salad for bible study ] [SL:DATE\_TIME on Friday ] ]

\bigskip

\textellipsis

\bigskip

\textbf{Example 19}\\
User: Remind me to make chicken dip for the watch party tomorrow.\\
Parse: [IN:CREATE\_REMINDER [SL:PERSON\_REMINDED me ] [SL:TODO make chicken dip for the watch party ] [SL:DATE\_TIME tomorrow ] ]

\bigskip

\textbf{Example 20}\\
User: Remind me to make the cookies for the bake sale.\\
Parse: [IN:CREATE\_REMINDER [SL:PERSON\_REMINDED me ] [SL:TODO make the cookies for the bake sale ] ]

\bigskip

\textbf{Query}\\
User: Remind me to make bars for the picnic on Sunday.\\
Parse:

\bigskip

\twocolumn

\newpage
\onecolumn

\newpage
\onecolumn
\subsection{Prompt Example for Spider}
\label{appendix:prompt_spider}

\noindent
Below are examples of database schema and text-to-SQL generation for Spider:

\bigskip
\textbf{/* Given the following database schema: */}\\
\texttt{CREATE TABLE IF NOT EXISTS "flight" ( "flno" text, "origin" text, "destination" text, "distance" text, "departure\_date" text, "arrival\_date" text, "price" text, "aid" text, PRIMARY KEY ("flno"), FOREIGN KEY ("aid") REFERENCES "aircraft"("aid") );}\\
\texttt{CREATE TABLE IF NOT EXISTS "aircraft" ( "aid" text, "name" text, "distance" text, PRIMARY KEY ("aid") );}\\
\texttt{CREATE TABLE IF NOT EXISTS "employee" ( "eid" text, "name" text, "salary" text, PRIMARY KEY ("eid") );}\\
\texttt{CREATE TABLE IF NOT EXISTS "certificate" ( "eid" text, "aid" text, PRIMARY KEY ("eid"), FOREIGN KEY ("aid") REFERENCES "aircraft"("aid"), FOREIGN KEY ("eid") REFERENCES "employee"("eid") );}

\smallskip
\textbf{/* Answer the following: How many employees do we have? */}\\
\textbf{SQL Query:} \texttt{SELECT count(*) FROM employee;}

\bigskip
\textbf{/* Given the following database schema: */}\\
\texttt{CREATE TABLE IF NOT EXISTS "Activity" ( "actid" text, "activity\_name" text, PRIMARY KEY ("actid") );}\\
\texttt{CREATE TABLE IF NOT EXISTS "Participates\_in" ( "stuid" text, "actid" text, FOREIGN KEY ("actid") REFERENCES "Activity"("actid"), FOREIGN KEY ("stuid") REFERENCES "Student"("StuID") );}\\
\texttt{CREATE TABLE IF NOT EXISTS "Faculty\_Participates\_in" ( "FacID" text, "actid" text, FOREIGN KEY ("actid") REFERENCES "Activity"("actid"), FOREIGN KEY ("FacID") REFERENCES "Faculty"("FacID") );}\\
\texttt{CREATE TABLE IF NOT EXISTS "Student" ( "StuID" text, "LName" text, "Fname" text, "Age" text, "Sex" text, "Major" text, "Advisor" text, "city\_code" text, PRIMARY KEY ("StuID") );}\\
\texttt{CREATE TABLE IF NOT EXISTS "Faculty" ( "FacID" text, "Lname" text, "Fname" text, "Rank" text, "Sex" text, "Phone" text, "Room" text, "Building" text, PRIMARY KEY ("FacID") );}

\smallskip
\textbf{/* Answer the following: How many faculty do we have? */}\\
\textbf{SQL Query:} \texttt{SELECT count(*) FROM Faculty;}

\bigskip

\textellipsis

\bigskip
\textbf{/* Given the following database schema: */}\\
\texttt{CREATE TABLE IF NOT EXISTS "artist" ( "Artist\_ID" text, "Name" text, "Country" text, "Year\_Join" text, "Age" text, PRIMARY KEY ("Artist\_ID") );}\\
\texttt{CREATE TABLE IF NOT EXISTS "exhibition" ( "Exhibition\_ID" text, "Year" text, "Theme" text, "Artist\_ID" text, "Ticket\_Price" text, PRIMARY KEY ("Exhibition\_ID"), FOREIGN KEY ("Artist\_ID") REFERENCES "artist"("Artist\_ID") );}\\
\texttt{CREATE TABLE IF NOT EXISTS "exhibition\_record" ( "Exhibition\_ID" text, "Date" text, "Attendance" text, PRIMARY KEY ("Exhibition\_ID"), FOREIGN KEY ("Exhibition\_ID") REFERENCES "exhibition"("Exhibition\_ID") );}

\smallskip
\textbf{/* Answer the following: How many artists do we have? */}\\
\textbf{SQL Query:} \texttt{SELECT count(*) FROM artist;}

\bigskip

\textbf{/* Given the following database schema: */}\\
\texttt{CREATE TABLE IF NOT EXISTS "stadium" ( "Stadium\_ID" text, "Location" text, "Name" text, "Capacity" text, "Highest" text, "Lowest" text, "Average" text, PRIMARY KEY ("Stadium\_ID") );}\\
\texttt{CREATE TABLE IF NOT EXISTS "singer" ( "Singer\_ID" text, "Name" text, "Country" text, "Song\_Name" text, "Song\_release\_year" text, "Age" text, "Is\_male" text, PRIMARY KEY ("Singer\_ID") );}\\
\texttt{CREATE TABLE IF NOT EXISTS "concert" ( "concert\_ID" text, "concert\_Name" text, "Theme" text, "Stadium\_ID" text, "Year" text, PRIMARY KEY ("concert\_ID"), FOREIGN KEY ("Stadium\_ID") REFERENCES "stadium"("Stadium\_ID") );}\\
\texttt{CREATE TABLE IF NOT EXISTS "singer\_in\_concert" ( "concert\_ID" text, "Singer\_ID" text, PRIMARY KEY ("concert\_ID"), FOREIGN KEY ("Singer\_ID") REFERENCES "singer"("Singer\_ID"), FOREIGN KEY ("concert\_ID") REFERENCES "concert"("concert\_ID") );}

\smallskip
\textbf{/* Answer the following: How many singers do we have? */}

\textbf{SQL Query:}

\newpage

\section{MLI Configuration}
\label{appendix:mli_configs}
Table~\ref{tab:mli_config} details the MLI configurations (linguistic property and injection intensity) applied to different exemplar selection methods for each dataset under various inference models. Configurations marked with “–” indicate settings that were not evaluated. All injections are applied at Layer 8 of the base retriever.

\begin{table*}[t]
  \centering
  \small
  \resizebox{\textwidth}{!}{
  \begin{tabular}{llcccc}
    \hline
    \textbf{Inference Model} & \textbf{Method + MLI} & \textbf{MTOP} & \textbf{SMCalFlow} & \textbf{TreeDST} & \textbf{Spider} \\
    \hline
    \multirow{3}{*}{Llama3-8B} 
      & STARE  & PT-5 & DEPS-4 & DEPS-1.5 & -- \\
      & BERT + MLI     & POS-5 & PT-1.5 & DEPS-1.5 & -- \\
      & EPR + MLI      & POS-6 & PT-4 & DEPS-4 & -- \\
    \hline
    \multirow{3}{*}{GPT-3.5-turbo} 
      & STARE  & -- & -- & -- & POS-5 \\
      & BERT + MLI     & -- & -- & -- & POS-5 \\
      & EPR + MLI      & -- & -- & -- & POS-5 \\
    \hline
    \multirow{3}{*}{GPT-4o-mini} 
      & STARE & POS-6 & PT-2.5 & PT-0.5 & POS-5 \\
      & BERT + MLI     & POS-5 & DEPS-3 & DEPS-1.5 & POS-5 \\
      & EPR + MLI      & DEPS-6 & DEPS-3 & DEPS-1.5 & POS-5 \\
    \hline
    \multirow{3}{*}{DeepSeek-V3} 
      & STARE & POS-4 & PT-2 & PT-3 & POS-6 \\
      & BERT + MLI     & POS-5 & PT-0.5 & PT-2 & POS-5 \\
      & EPR + MLI      & DEPS-6 & PT-2 & PT-2 & POS-5 \\
    \hline
  \end{tabular}
  }
  \caption{MLI configurations (property - intensity) for each inference model across datasets. “--” indicates the method is not evaluated for that task-model pair.}
  \label{tab:mli_config}
\end{table*}

\end{document}